\pdfoutput=1

\documentclass[11pt]{article}

\usepackage[preprint]{tmlr}

\usepackage{times}
\usepackage{latexsym}

\usepackage[T1]{fontenc}

\usepackage[utf8]{inputenc}

\usepackage{hyperref}

\definecolor{ruddy}{rgb}{1.0, 0.0, 0.16}
\definecolor{gblue}{RGB}{29, 144, 255}
\definecolor{royalblue}{rgb}{0.25, 0.41, 0.88}

\hypersetup{
  colorlinks,
  citecolor=royalblue,
  linkcolor=ruddy,
  urlcolor=royalblue}

\usepackage{microtype}
\usepackage{bbm}
\usepackage{amsmath}
\usepackage{graphicx}
\usepackage{hyperref}
\usepackage{authblk}
\newcommand{\dapt}[0]{DAPT }
\usepackage{lipsum, enumitem}
\title{Improving Few-shot Generalization of Safety Classifiers via Data Augmented Parameter-Efficient Fine-Tuning}

\author[1]{\bf Ananth Balashankar}
\author[1]{\bf Xiao Ma}
\author[1]{\bf Aradhana Sinha}

\author[1]{\bf Ahmad Beirami}
\author[1]{\\\bf Yao Qin}
\author[1]{\bf Jilin Chen}
\author[2]{\bf Alex Beutel\thanks{Work done while author was at Google.}\ \ }
\affil[1]{Google Research}
\affil[2]{OpenAI}

\begin{document}
\maketitle
\begin{abstract}
As large language models (LLMs) are widely adopted, new safety issues and policies emerge, to which existing safety classifiers do not generalize well.  If we have only observed a few examples of violations of a new safety rule, how can we build a classifier to detect violations?
In this paper, we study the novel setting of domain-generalized few-shot learning for LLM-based text safety classifiers.
Unlike prior few-shot work, these new safety issues can be hard to uncover and we do not get to choose the few examples. We demonstrate that existing few-shot techniques do not perform well in this setting, and rather we propose to do parameter-efficient fine-tuning (PEFT) combined with augmenting training data based on similar examples in prior existing rules.
We empirically show that our approach of similarity-based data-augmentation + prompt-tuning (DAPT) consistently outperforms baselines that either do not rely on data augmentation or on PEFT by 7-17\% F1 score in the Social Chemistry moral judgement and 9-13\% AUC in the Toxicity detection tasks, even when the new rule is loosely correlated with existing ones.

\end{abstract}

\section{Introduction}

Safety classifiers are important tools for limiting potential harms \citep{toxicity2018,lees2021capturing} introduced by large language models \citep{mozes2023agile}, but they often don't generalize well due to limited data and continuously emerging safety risks.
Traditionally, safety classifiers are trained with supervised fine-tuning, which requires thousands of examples for each safety rule or policy.
In a real-world setting, however, collecting large amounts of data can be costly.
Further, new safety risks emerge quickly -- a news event, new slur, or attack pattern, requiring safety classifiers to generalize to new safety rules based on a few examples.

To solve this domain-generalized few-shot problem in safety classifiers, in-context learning (ICL) \citep{brown2020language, cot} and PEFT methods \citep{lester-etal-2021-power, li-liang-2021-prefix} are promising directions. But it is unclear to what extent these methods will result in safety classifiers that: (1) generalize well to a new safety rule with only a few examples \citep{wang-etal-2022-measure}; and (2) perform well agnostic to the choice of the few examples.
In this paper, we first show that ICL and PEFT methods alone are not sufficient to result in few-shot generalization of safety classifiers.
Specifically, ICL does not generalize over a random choice of few-shot examples as most prior work in few-shot prompting  typically hand-craft the few-shot examples and is  infeasible to be used with training data due to large sequences. On the other hand, PEFT methods such as prompt tuning  have the advantage that they can be tuned on a given training dataset with no manual prompt engineering but do not generalize with only a few examples ($\approx$ 5; they require about $\approx$ 100) on benchmark safety datasets \citep{mozes2023agile}. 

To overcome these limitations, we propose \dapt, a method that combines insights from contextual \underline{D}ata \underline{A}ugmentation \citep{arthaud-etal-2021-shot} and domain adaptation \citep{NIPS2017_21c5bba1} --  augmenting examples from existing safety rules that are most similar to few-shot examples from the new safety rule as training data for \underline{P}rompt-\underline{T}uning, a PEFT method \citep{lester-etal-2021-power}. We compare \dapt to random data augmentation and fine-tuning methods. LLM-based synthetic data generation methods are out of scope in our study \citep{bai2022constitutional}. We find that by augmenting similar examples, we learn prompt-tuned safety classifiers that generalize to new safety rules, even when that rule is loosely correlated with existing rules.

To summarize, our key contributions include: 
\begin{enumerate}[leftmargin=*]
\itemsep 0em 
    \item \textbf{Problem Formulation}: We study how best to make safety classifiers robust to new types of safety concerns. Domain-generalized few-shot learning has not been studied before in LLM-based safety classifiers, and existing few-shot techniques like ICL or PEFT or simple data augmentation alone do not perform well. 
    \item \textbf{Method}: We propose a prompt-tuning based method that augments examples from the existing fine-tuning data that are most similar to the few-shot examples, and demonstrate that it outperforms baselines on 2 benchmark safety tasks by 7-17\% in F1 score and 9-13\% AUC.
    \item \textbf{Empirical insights}: We discuss implications for future safety classifier research and real-world use by analyzing the performance gains based on the choice of 5-shot examples, and the augmented data.
\end{enumerate}



\section{Problem Formulation}

We look at a safety classifier which rates whether an action was safe given a certain context. 
For example, in the Social Chemistry dataset, in the context, ``\textit{ Man's Snoring Threatens to Chase Wife out of Marriage}'', the safety classifier predicts how good/bad is it to do this action morally - ``\textit{preventing someone from being able to sleep.}'' The safety label (bad) assigned to the input can vary on a 5-point Likert scale from very-bad to very-good \cite{likert1932technique}.
As notations, we refer to the safety classifier as $f$ which takes as input $\mathbf{x}$ (encapsulates both context and action), and predicts the output class $f(\mathbf{x})$. 

Further, we assume that the train and test data of this classifier is composed by a set of safety rules $R$ that govern the crowd-sourcing guidelines (e.g. violence, hate speech, ethical, medicine guidelines). Each input $\mathbf{x}$ is labeled $y$, and belongs to one or more rules $r_i \in R = \{r_1, r_2, r_3, .., r_n\}$. Traditionally, the classifier is evaluated using a metric $A$ (e.g. accuracy, AUC, F1) on the entire test set. To measure few-shot generalization, we use sliced metrics over test sets for each safety rule $r_i$ given by $A_i$. 

Since new social trends might emerge or safety policies might evolve, we assume that a new safety rule $r_{n+1}$ emerges. We assume that there is sufficient training data to fine-tune on the first $n$ safety rules and only a few ($k$) examples available that belong to the $n+1^{th}$ safety rule. In the rest of the paper, we will focus on how to train a classifier that uses the $k$ few-shot examples from the $n+1^{th}$ rule to maximize $A_{n+1}$. We compare our method's performance to the out-of-distribution (OOD) performance of fine-tuned classifier $f$ and competing few-shot learning methods that use $f$ as the base model. Our safety-inspired formulation is a novel few-shot variant of domain generalization \citep{muandet2013domain,gulrajani2021in}: \emph{Given a random sample of $k$ examples from the rule: $r_{n+1}$, can we train $f'$ such that $A'_{n+1} > A_{n+1}$?}

\section{Related Work}

Past work in LLM few-shot learning have found success with parameter efficient finetuning (PEFT) methods that only change a small fraction of the LLM's parameters training typically on a few hundred examples. The tuned parameters can either belong in the input embedding (i.e. prompt or prefix tuning) \citep{lester-etal-2021-power, li-liang-2021-prefix}, the output layer (i.e. adapters) \citep{lora}, or intermediate layers \citep{liu2022few}.
PEFT methods are often compared with in-context learning (ICL), which uses natural language prompts as instructions, often providing few-shot examples in natural text form \citep{brown2020language}. Past work to improve ICL has manually engineered adversarially robust prompts \citep{raman2023modeltuning}, extended reasoning based prompts using chain-of-thought \citep{cot}, self-consistency \citep{wang2023selfconsistency}, recitation \citep{sun2023recitationaugmented}, self-ask \citep{press2022measuring}, multi-step \citep{lewkowycz2022solving}, least-to-most \citep{zhou2023leasttomost} zero-shot or few-shot exemplars.

More broadly, prior work in domain adaptation has studied how to adapt classifiers trained on a source domain to a target domain.
Domain adaptation methods typically seek to learn a better domain-invariant representation \citep{nishida2020unsupervised, karouzos-etal-2021-udalm}. 
In a few-shot setting, past work has done this by using a domain classifier in an adversarial setting \citep{NIPS2017_21c5bba1, wang-etal-2019-adversarial}. 
Data augmentation has also been used in the few-shot setting. When domain knowledge is available (in our paper it is not), domain structure can be used to do template-based or scoring-based data augmentation \citep{oguz-vu-2021-shot,ma-etal-2019-domain}. Domain-agnostic
techniques include compute-expensive example generation methods \citep{hong2018conditional, fabbri-etal-2021-improving, zhou-etal-2022-flipda}, and cheap source domain data re-weighting and augmentation \citep{jiang-zhai-2007-instance, kumar2019closer}. Since we expect to adapt to new safety rules very frequently in our problem formulation, we focus on the cheaper domain-agnostic source domain data augmentation methods.  

Our formulation of few-shot generalization has been referred to as "domain generalized" or "meta" few-shot learning, previously studied in image recognition \citep{liang2021boosting,meta_reg_network} and text keyword extraction tasks \citep{priyanshu2022adaptkeybert}. 
In an unlabeled target domain setting, this has been studied as continual few-shot learning, using complex methods to build domain-invariant representations \citep{ke-etal-2022-continual}.
In the closely related model-agnostic meta-learning framework, classifiers also train a base model  on few-shot examples in the target domain \citep{han2021metalearning}, and focus on making the base model readily adaptable to downstream tasks \citep{sharaf-etal-2020-meta, bansal-etal-2020-learning, sui-etal-2021-knowledge}. LLMs are good meta-learners \citep{radford2019language}, and so we take inspiration from meta-learning techniques that augment data from the source domain that are most useful to the target domain \citep{jain2023data}.


\section{Methodology}
We propose our method, \dapt -- that combines data augmentation and prompt tuning to improve few-shot generalization of safety classifiers. Specifically, we use simple cost-efficient data augmentation techniques \citep{zhou-etal-2022-flipda} to expand from our five examples from the new safety rule to $k=100$ typically required for successful prompt-tuning methods. Costly generative methods are infeasible in our problem setting that requires such frequent domain adaptation.
Instead, we augment data by selecting $k=100$ examples from existing safety rules (source) that are most similar to our five examples from the new safety rule (target). We test \dapt on three axes of similarity for data augmentation:

\begin{itemize}
    \setlength{\itemsep}{0pt}
    \setlength{\parskip}{0pt}
    \item \underline{Cosine} similarity of bag-of-words of the source and target sentences after removing stop-words \citep{harris1954distributional}.
    \item \underline{ReCross}, an unsupervised retrieval-ranking method based on similarity between source and target sentences' SentenceBERT embeddings \citep{lin2022unsupervised}.
    \item Contextual data augmentation (\underline{CDA}), where we find source sentences that have similar context as that of the diff in the target sentence \citep{arthaud-etal-2021-shot}.
\end{itemize}

We choose prompt tuning \citep{lester-etal-2021-power}, a PEFT method, in \dapt and not ICL for two reasons. First, compared to ICL, PEFT methods are known to be more accurate and less costly \citep{liu2022few}-- a nontrivial advantage in a problem set-up where we will have to repeatedly adapt to new rules. Second, ICL relies on few-shot examples manually chosen for the task/model through repeated trials; ICL is not designed to work on a random sample of few-shot examples. This constraint makes it difficult to use ICL in a production use case for a safety task. Few shot examples for safety often arise from real-world user interactions or red-teaming adversarial efforts -- which are inherently random processes.

\vspace{-.1in}
\section{Evaluation}
\vspace{-.07in}
We choose the following 2 safety tasks based on social norms, crucial to mitigating societal harms. We present a summary of the evaluation below, with more details in Appendix.

\textbf{Social Chemistry 101}:
In this dataset, we study a 5-class classification task to identify if a specific action is morally judged to be appropriate given a situation \citep{forbes-etal-2020-social}. The dataset, sourced from online forums, is split into 5 moral foundation rules: care/harm, fairness/cheating, loyalty/betrayal, authority/subversion, sanctity/degradation. We evaluate each safety rule in a hold-one-out strategy -- the base safety classifier is trained on fine-tuning data from 4 safety rules and we evaluate 5-shot generalization on the 5$^{th}$ rule.

\textbf{Toxicity:}
The toxicity detection task is a binary classification task where Wikipedia comments are annotated by human raters \citep{toxicity2018}. We split the data across 5 safety rules: toxic, obscene, threat, insult, and identity-hate.

\textbf{Baselines:}
LLM based safety classifiers have demonstrated few-shot capabilities \citep{mozes2023agile}, and so we use the 62B PaLM model \citep{chowdhery2022palm} fine-tuned on our 2 safety tasks as the base model.
In addition to the base classifier, we evaluate against fine-tuning \citep{howard2018universal}, prompt-tuning \citep{lester-etal-2021-power}, in-context few-shot \citep{brown2020language}, low-rank adapters (LoRA) \citep{lora}, and chain-of-thought \citep{cot} (using rule-of-thumb in social chemistry), and transfer learning \citep{gupta-etal-2020-effective} baselines. We also evaluate against automated prompt generation \citep{gao-etal-2021-making} to compare against gradient-based natural language prompt learning techniques. Further, to understand the impact of the data augmented, we evaluate against a method that samples 100 random examples from the existing safety rules, and to measure the value of using PEFT, we evaluate against supervised fine-tuning (SFT) with the 3 data-augmentation methods . 

\vspace{-.1in}
\section{Results}
\vspace{-.07in}

\textbf{Summary}: We evaluate on 5 new safety rules in each of the 2 safety tasks, and find that \dapt outperforms the base OOD, few-shot, and data-augmentation only classifiers. In Tables \ref{social} and \ref{toxicity}, each of the columns report generalization performance over the test set from the new safety rule.

\textbf{Baseline few-shot methods do not generalize}: We observe that none of the methods in the 5-shot setting improve upon the OOD performance of the base model. The base model is fine-tuned on all other safety rules' data in a leave-one-out manner, and baseline methods in the 5-shot report metrics over 5 random examples from the new safety rule.

\textbf{PEFT methods generalize better than SFT methods with data augmentation}: We see that \dapt methods outperform fine-tuning (SFT) counterparts, which shows the value of using parameter efficient methods in overly limited data settings. We argue that fine-tuning all the 62B parameters over 105 examples in SFT is sub-optimal, as compared to 50K parameters in \dapt leading to better generalization. Compared to the 5-shot setting, methods that are further augmented with 100 examples from existing safety rules, regardless of the similarity criteria - highlighting the value of similar data augmentation. Further, we see that random data augmentation does not improve prompt tuning method's performance.

\begin{table}[t!]
\centering
\resizebox{0.8\columnwidth}{!}{
\begin{tabular}{llllll}
\textbf{Held-out safety rule $\rightarrow$} & \textbf{Harm} & \textbf{Fairness} & \textbf{Betrayal} & \textbf{Degrade} & \textbf{Authority} \\\hline
Base (OOD) & 31.0 & 24.5 & 23.2 & 16.5 & 19.2 \\\hline
\multicolumn{6}{c}{5-shot}\\\hline
Fine-tune (SFT) & 31.1$_{\pm 1.1}$ & 25.2$_{\pm 1.6}$ & 23.4$_{\pm 2.0}$ & 18.1$_{\pm 3.1}$ & 19.3$_{\pm 1.5}$ \\
Prompt-Tune (PT) & 31.6$_{\pm 0.9}$ & 25.3$_{\pm 1.0}$ & 23.7$_{\pm 2.1}$ & 16.8$_{\pm 2.8}$ & 19.8$_{\pm 1.3}$ \\
ICL few-shot & 31.9$_{\pm 1.3}$ & 25.5$_{\pm 2.0}$ & 23.8$_{\pm 1.7}$ & 18.3$_{\pm 3.2}$ & 20.0$_{\pm 2.0}$ \\
Chain-of-thought & 31.9$_{\pm 1.2}$ & 25.9$_{\pm 1.6}$ & 24.2$_{\pm 1.9}$ & 18.5$_{\pm 4.0}$ & 19.9$_{\pm 1.5}$ \\
{LoRA} & 32.1$_{\pm 1.0}$ & 26.1$_{\pm 1.5}$ & 25.5$_{\pm 0.9}$ & 22.1$_{\pm 1.9}$ & 21.3$_{\pm 1.3}$ \\
Automated prompt generation & 32.0$_{\pm 1.1}$ & 25.6$_{\pm 1.5}$ & 23.9$_{\pm 1.3}$ & 18.6$_{\pm 2.1}$ & 20.4$_{\pm 1.3}$\\
Transfer learning & 31.9$_{\pm 1.9}$ & 25.9$_{\pm 1.4}$ & 24.7$_{\pm 1.8}$ & 19.9$_{\pm 2.0}$ & 19.5$_{\pm 1.7}$ \\\hline
\multicolumn{6}{c}{5-shot + Data Augmentation (DA: 100 examples)}\\\hline
SFT + CDA & 32.9$_{\pm 0.8}$ & 32.6$_{\pm 0.9}$ & 26.9$_{\pm 1.9}$ & 26.1$_{\pm 0.9}$ & 24.7$_{\pm 1.0}$\\
SFT + Cosine & 35.9$_{\pm 0.7}$ & 34.5$_{\pm 1.7}$ & 29.2$_{\pm 2.1}$ & 28.5$_{\pm 1.2}$ & 29.3$_{\pm 1.2}$\\
SFT + ReCross & 36.1$_{\pm 0.7}$ & 34.6$_{\pm 1.7}$ & 29.6$_{\pm 2.1}$ & 28.2$_{\pm 1.4}$ & 29.2$_{\pm 1.6}$\\
\dapt (random) & 31.5$_{\pm 1.5}$ & 25.3$_{\pm 1.6}$ & 24.6$_{\pm 2.3}$ & 19.6$_{\pm 2.5}$ & 20.1$_{\pm 2.4}$ \\
\dapt (CDA) & 36.3$_{\pm 0.7}$ & 34.8$_{\pm 0.8}$ & 33.4$_{\pm 1.0}$ & 35.5$_{\pm 1.2}$ & 34.2$_{\pm 1.3}$ \\
\dapt (Cosine) & \textbf{38.4$_{\pm 1.1}$} & \textbf{37.3$_{\pm 0.9}$} & \textbf{36.8$_{\pm 1.5}$} & \textbf{38.2$_{\pm 0.9}$} & \textbf{36.1$_{\pm 1.0}$} \\
\dapt (ReCross) & \textbf{38.1$_{\pm 0.9}$} & \textbf{37.2$_{\pm 0.8}$} & \textbf{36.6$_{\pm 1.2}$} & \textbf{38.0$_{\pm 0.7}$} & \textbf{36.3$_{\pm 0.7}$}
\end{tabular}
}
\caption{Improvement in average F1 (\%) by 7-17\% in Social Chemistry 101 task using our \dapt methods. ($\pm$ 5-shot standard error on 5 random samples) }
\label{social}
\end{table}

\begin{table}[ht!]
\centering
\resizebox{0.8\columnwidth}{!}{
\begin{tabular}{llllll}

\textbf{Held-out safety rule $\rightarrow$}    & \textbf{Toxic}       & \textbf{Obscene}                   & \textbf{Threat}      & \textbf{Insult}      & \textbf{Hate} \\\hline
Base (OOD)                     & 0.72                 & { 0.78}        & 0.71                 & 0.69                 & 0.73                   \\\hline
\multicolumn{6}{c}{5-shot}\\\hline
Fine-tune (SFT)                   & 0.72$_{\pm 0.02}$          & { 0.77$_{\pm 0.02}$} & 0.71$_{\pm 0.03}$          & 0.70$_{\pm 0.03}$          & 0.72$_{\pm 0.05}$            \\
Prompt-tune (PT)                  & 0.80$_{\pm 0.03}$          & { 0.82$_{\pm 0.03}$} & 0.76$_{\pm 0.02}$          & 0.74$_{\pm 0.01}$          & 0.75$_{\pm 0.01}$            \\
ICL few-shot                 & 0.78$_{\pm 0.01}$          & 0.79$_{\pm 0.04}$                        & 0.73$_{\pm 0.06}$          & 0.72$_{\pm 0.05}$          & 0.72$_{\pm 0.04}$            \\
LoRA                         & 0.79$_{\pm 0.03}$          & 0.83$_{\pm 0.03}$                        & 0.78$_{\pm 0.01}$          & 0.78$_{\pm 0.03}$          & 0.76$_{\pm 0.01}$            \\
Automated prompt generation &  0.79$_{\pm 0.03}$ & 0.80$_{\pm 0.02}$ & 0.75$_{\pm 0.04}$ &  0.74$_{\pm 0.04}$ & 0.73$_{\pm 0.02}$ \\
Transfer learning            & 0.72$_{\pm 0.02}$          & 0.78$_{\pm 0.01}$                        & 0.72$_{\pm 0.04}$          & 0.75$_{\pm 0.04}$          & 0.75$_{\pm 0.02}$            \\\hline
\multicolumn{6}{c}{5-shot + Data Augmentation (DA: 100 examples)}\\\hline
SFT + CDA & 0.80$_{\pm 0.01}$          & 0.84$_{\pm 0.02}$                        & 0.80$_{\pm 0.01}$          & 0.78$_{\pm 0.02}$          & 0.75$_{\pm 0.02}$            \\
SFT + cosine & 0.83$_{\pm 0.02}$          & 0.84$_{\pm 0.01}$                        & 0.81$_{\pm 0.02}$          & 0.82$_{\pm 0.01}$          & 0.82$_{\pm 0.02}$   \\
SFT + ReCross & 0.82$_{\pm 0.01}$          & 0.84$_{\pm 0.01}$                        & 0.82$_{\pm 0.01}$          & 0.83$_{\pm 0.01}$          & 0.81$_{\pm 0.01}$ \\
\dapt (CDA) & 0.84$_{\pm 0.03}$          & 0.86$_{\pm 0.01}$                        & 0.84$_{\pm 0.03}$          & 0.79$_{\pm 0.04}$          & 0.77$_{\pm 0.02}$            \\
\dapt (Cosine)            & \textbf{0.89$_{\pm 0.01}$} & \textbf{0.92$_{\pm 0.01}$}               & \textbf{0.89$_{\pm 0.02}$} & \textbf{0.87$_{\pm 0.03}$} & \textbf{0.86$_{\pm 0.02}$}   \\
\dapt (ReCross)  & \textbf{0.90$_{\pm 0.02}$} & \textbf{0.90$_{\pm 0.03}$}               & \textbf{0.91$_{\pm 0.01}$} & \textbf{0.88$_{\pm 0.01}$} & \textbf{0.89$_{\pm 0.04}$}  
\end{tabular}
}
\caption{Improvement in AUC of Toxicity detection task by 9-13\% using our \dapt methods ($\pm$ 5-shot standard error on 5 random samples)}
\label{toxicity}
\end{table}

\textbf{Augmenting with similar examples consistently helps, even given worst case few-shot examples}: We find that cosine and ReCross similarity outperforms contextual and random DA. Further, we find that the base PT method is brittle and quite sensitive to the choice of 5 support examples: picking target examples that are furthest from each other, and closest to source examples perform much worse. Our method is not brittle: it performs equally well even under worst-case choices for the five support examples (Table \ref{settings} in Appendix). Our method's performance variance is also much lower, which is indicative of more robust generalization.
Finally, we demonstrate AUC gains even in safety rules with low cross-correlation with existing rules (Table \ref{corr}), showing that our DA methods find the most relevant examples (Table \ref{samples}) and do not entirely rely on distributional overlap to improve generalization. We varied the number of examples augmented in \dapt and found diminishing gains beyond 100 examples (Table \ref{size}).




\vspace{-.1in}
\section{Discussion \& Conclusion}
\vspace{-.07in}
We demonstrate that existing safety classifiers do not generalize to new safety rules in a few-shot setting on 2 benchmark safety classification tasks.
We believe this is critical for safety classifiers that are characterized by unique challenges.
Unlike math reasoning problems that may have stable answers overtime, new safety risks emerge: e.g., a news event, a new slur, a new attack pattern.
When this happens, often we only have a few examples available, and we show that existing approaches are not sufficient to quickly mitigate the new safety risks.

We propose \dapt - a robust data-augmented prompt-tuning method using only 5 random few-shot examples from a new safety rule and augmenting 100 similar examples from existing safety rules. 
Our proposed approach of augmenting the examples is practical and improves generalization to new safety rules by 7-17\% F1 score on the Social Chemistry 101 moral judgement, and by 9-13\% AUC on the Jigsaw Toxicity detection tasks.
We believe that this approach is a valuable building block toward a framework that continually accommodates a large number of emergent rules without regressing classifier performance on prior data.

\section*{Limitations}
Our method shows improvement in few-shot performance of safety classifiers using data from the existing dataset.
Future work can explore synthetic data augmentation that uses the 5-shot examples to generate noisy data and compare its effectiveness \citep{bai2022constitutional}. The splits we generate in safety classification are provided in the benchmark datasets, and may have non-trivial correlation across safety rules (e.g. correlation between toxic and obscene is 0.68) and hence the gains may not be generalizable to a non-safety task with lower correlation across rules. Our evaluation is limited to 5-shot generalization on the PaLM 62B LLM which has been shown to outperform T5-XXL \citep{mozes2023agile} and GPT-3 \citep{chowdhery2022palm} on benchmark safety and natural language understanding tasks. The training and evaluation datasets we evaluate over are in English, and is known to be from human annotators, majority of whom are from the U.S and identify as white \citep{forbes-etal-2020-social}. The few-shot generalization of safety classifiers we demonstrate is grounded in notions of safety and social norms prevalent in these demographic groups, and studying how they extend to other locales, demographics, and languages is left as future work.

\section*{Ethics Statement}
Safety is a highly contextualized concept based on social norms and beliefs. Learning a classifier that broadly categorizes this nuanced concept into rules or policies, though imperfect, is unavoidable in real-world NLP applications. To this end, we have evaluated on one of the datasets that belong to a body of recent work that contextualizes social bias \citep{sap-etal-2020-social}, ethical norm reasoning \citep{forbes-etal-2020-social}, and reasoning about harms \citep{zhou2023cobra}. We believe, orthgonally, safety classifiers need to be more generalizable both from a domain-specific manner using all available context, and in a domain-agnostic manner leveraging insights from data augmentation and parameter efficient fine-tuning. Though our work focuses on the latter, we acknowledge that combining contextual and reasoning based approaches with our methods can further improve safety of NLP applications.
\bibliography{anthology,custom}
\bibliographystyle{tmlr}

\clearpage
\appendix
\section{Appendix}

\setcounter{table}{0}
\renewcommand{\thetable}{A\arabic{table}}
\textbf{Implementation details:} 
\label{impl}
The Social Chemistry and Toxicity datasets were processed using Tensorflow Seqio library. We use the Pathways framework for training and evaluating the few-shot robustness. We use the following hyper-parameters across our methods: batch size: 16 (we upsample from 5 when we have less than 16 examples), learning rate: 0.3 for PEFT methods, and 0.005 for FT methods (chosen based on validation split accuracy), few-shot number of steps: 200, the baseline model was finetuned for 10,000 steps from the base PaLM 62B checkpoint. The prompt embeddings are in 50 in length, with each having a width of 1024 units (50K parameters). They are initialized using sampling from 5000 most common words in T5 sentence-piece vocabulary as given in \citet{lester-etal-2021-power}. 

\textbf{Social Chemistry:}
We pose the following question to the safety classifier: "In the situation - <situation>, how good/bad is it to do this action morally: <action>? (a) very-bad (b) bad (c) ok (d) good (e) very-good" and post-process the response as a multiple-choice answer \citep{cot}, and we report per-rule average F1 score. Since this task was originally proposed as a text generation task, we confirmed that a classifier that uses training data from all 5 safety rules has an F1 score of 53.2\%.

To train the base model, the available training data in the Social Chemistry dataset was sampled to 15,000 examples from each safety rule for equal comparison and not to allow a rule with large training data to dominate others (e.g. care/harm: 103K, fairness/cheating: 38K, loyalty/betrayal: 42K, authority/subversion: 23K , sanctity/degradation: 16K).

\textbf{Toxicity:}
In the toxicity detection task, all examples are labeled on all 5 safety rules, however the positive examples that belong to the new rule are held-out when we trained the base model for each new rule in a  leave-one-out manner. Hence, during fine-tuning of the base model, we chose a sample of 100K out of the total 159K available. The number of positive examples vary by safety rule in the training data as follows: (toxic: 15.2K , hate: 1.4K, obscene: 8.4K, threat: 0.4K, insult: 7.8K). 

We pose the following question to the classifier: "Is this <comment> <rule>? (a) yes (b) no", and report the per-rule AUC. Also, there is no context mentioned in the toxicity dataset, and hence we omit it in the question posed, assuming the toxicity of the comment in all contexts is asked.
The toxicity dataset has another safety rule called ``severe toxic'' which we ignore as it meant as a subset of toxic, and does not fit our new safety rule formulation. 

\textbf{In-distribution accuracy:}
The in-distribution F1 score of the base classifier in the Social Chemistry dataset was 53.2\% and the AUC of the base classifier in the toxicity detection task was 0.81. Since our DAPT classifiers are not meant to be used on the in-distribution dataset, we did not present the results in the main paper, but with DAPT these results vary between 49.9-54.1 F1 score and 0.78-0.82 AUC in both the tasks respectively.

\begin{table*}[h]
\centering
\resizebox{0.8\linewidth}{!}{
\begin{tabular}{llllll}
\textbf{Held out safety rule $\rightarrow$} & \textbf{Harm} & \textbf{Fairness} & \textbf{Betrayal} & \textbf{Degrade} & \textbf{Authority} \\\hline
Base & 31.0 & 24.5 & 23.2 & 16.5 & 19.2 \\
Prompt-tune (PT): 5 random  & 31.6$_{\pm 0.9}$ & 25.3$_{\pm 1.0}$ & 23.7$_{\pm 2.1}$ & 16.8$_{\pm 2.8}$ & 19.8$_{\pm 1.3}$               \\
$\hookrightarrow$ + cosine DA & \textbf{38.4$_{\pm 1.1}$} & \textbf{37.3$_{\pm 0.9}$} & \textbf{36.8$_{\pm 1.5}$} & \textbf{38.2$_{\pm 0.9}$} & \textbf{36.1$_{\pm 1.0}$} \\\hline
\#1: PT: Closest 5 within target class & 20.0 & 23.5 & 27.2 & 21.6 & 23.3\\
$\hookrightarrow$ + cosine DA & 37.5 & 36.5 & 36.2 & 37.9 & 35.4\\
\hline
\#2: PT: Furthest 5 within target class & 16.7 & 14.0 & 25.9 & 18.3 & 20.2\\
$\hookrightarrow$ + cosine DA & 35.6 & 35.1 & 34.9 & 35.8 & 33.9\\
\hline
\#3: PT: 5 closest to source classes & 9.0 & 17.7 & 15.1 & 14.2 & 16.3\\
$\hookrightarrow$ + cosine DA & 32.7 & 34.9 & 33.7 & 33.5 & 35.2\\
\hline
\#4: PT: 5 furthest to source classes & 15.2 & 25.1 & 29.5 & 17.6 & 23.7\\
$\hookrightarrow$ + cosine DA & 33.7 & 35.3 & 34.3 & 34.1 & 35.7\\
\end{tabular}
}
\caption{Average F1 score when 5-shot examples are chosen based on BERT-Large embedding distance to other examples. Computing the subset with the max/min intra-subset distance within the target class is NP-Hard \citep{GHOSH1996175}, so we use the Drop-Add Simple Tabu Search approximation algorithm  \citep{approx-algo} for \#1 and \#2.
The closest and furthest support examples from the source data are computed exactly (\#3, \#4). The table shows that the base PT method is brittle and quite sensitive to the choice of five support examples: performance is better when the support examples are closer to each other (\#1 > \#2), or further from the source domain examples (\#4 > \#3). Our proposed approach, on the other hand, is not sensitive to the choice of few-shot support examples. Its performance variance is much lower, and  its F1 scores are equally high across all settings in Table \ref{social}.}
\label{settings}
\end{table*}

\begin{table*}[h]
\centering
\resizebox{0.6\columnwidth}{!}{
\begin{tabular}{
l 
r 
r 
r 
r 
r }
\textbf{}                               & \textbf{toxic} &  \textbf{obscene} &  \textbf{threat} &  \textbf{insult} & \textbf{hate}\\\hline
{\textbf{toxic}}   & {1.00}                                                       & {0.68}                                                         & {0.16}                                                        & {0.65}                                                        & {0.27}                                                      \\
{\textbf{obscene}} & {0.68}                                                       & {1.00}                                                         & {0.14}                                                        & {0.74}                                                        & {0.29}                                                      \\
{\textbf{threat}}  & {0.16}                                                       & {0.14}                                                         & {1.00}                                                        & {0.15}                                                        & {0.12}                                                      \\
{\textbf{insult}}  & {0.65}                                                       & {0.74}                                                         & {0.15}                                                        & {1.00}                                                        & {0.34}                                                      \\
{\textbf{hate}}    & {0.27}                                                       & {0.29}                                                         & {0.12}                                                        & {0.34}                                                        & {1.00}                                                     
\end{tabular}
}
\caption{Pearson correlation coefficient between the various safety rules in the Toxicity dataset. In Table \ref{toxicity}, we see improvement in AUC on both rules with high pearson correlation coefficient when held-out from the rest of the safety rules: (toxic: 0.74, obscene: 0.68, insult: 0.66) and those with low correlation with the rest: (threat: 0.15, hate: 0.27)}
\label{corr}
\end{table*}

\begin{table*}[]
\centering
\resizebox{0.9\linewidth}{!}{
\begin{tabular}{p{12cm}|l}
\textbf{Input} & \textbf{Label}\\\hline
\multicolumn{2}{c}{5-shot examples from "Care" safety rule in Social Chemistry moral judgment task}\\\hline
In the situation - Man's Snoring Threatens to Chase Wife out of Marriage, how good/bad is it to do this action morally - preventing someone from being able to sleep.? & bad \\
In the situation - He would have little to no physical activity, he ate and sat around., how good/bad is it to do this action morally - keeping in shape.? & good  \\
{In the situation - Mom's Downward Spiral Makes Daughter Fear Possible Crash, how good/bad is it to do this action morally - being worried about your parents.?} & ok  \\
In the situation - attempting to help calm down a child who isint mine, how good/bad is it to do this action morally - caring about the well-being of others.? & very good  \\
In the situation - Best Friend Vandalizes Rival Classmate's House, how good/bad is it to do this action morally - vandalizing the property of others? & very bad \\\hline
\multicolumn{2}{c}{Top 10 examples from 4 other existing safety rules that are augmented}\\\hline
In the situation - Wife Is Embarrassed by Man's Need for Applause, how good/bad is it to do this action morally - people applauding for someone who did a good job.? & ok \\
In the situation - Change in Man's Personality When Drinking Threatens Wife, how good/bad is it to do this action morally - threatening your partner.? & bad  \\
In the situation - blocking someone over a roleplay, how good/bad is it to do this action morally - blocking someone's bad behavior over a roleplay.? & ok \\
In the situation - It's too bad that she's..., how good/bad is it to do this action morally - expecting someone to be perfect.? & bad  \\
In the situation - Friendship Marred by Man's Harsh Treatment of His Wife, how good/bad is it to do this action morally - ending a friendship when someone turns out to be a bad person.? & ok \\
In the situation - Marriage to Man Doesn't End Wife's Love for Another Woman, how good/bad is it to do this action morally - marrying someone when you are in love with someone else.? & bad  \\
In the situation - Change in Man's Personality When Drinking Threatens Wife, how good/bad is it to do this action morally - calling the police on someone who is drunk and threatening people.? & good \\
In the situation - Man's Attachment To Family Threatens To Derail Marriage Plans, how good/bad is it to do this action morally - caring about your family? & good \\
In the situation - Man Fears Wife's Revenge if He Ends Abusive Marriage, how good/bad is it to do this action morally - abusing your partner? & bad \\
In the situation - wasting my dad's time, how good/bad is it to do this action morally - taking someone's time from them for no good reason.?  & bad\\
\end{tabular}
}
\caption{Examples of 5-shot examples from the care/harm safety rule in Social Chemistry moral judgement task, and a sample of the top 10 most similar examples augmented based on cosine similarity in \dapt. The labels are on a 5-point Likert scale ranging from very bad to very good.}
\label{samples}
\end{table*}

\begin{table*}[ht]
\centering
\resizebox{0.8\linewidth}{!}{
\begin{tabular}{llllll}
\textbf{DA Size / Held-out Rule} & \textbf{Harm} & \textbf{Fairness} & \textbf{Betrayal} & \textbf{Degrade} & \textbf{Authority} \\\hline
Base & 31.0 & 24.5 & 23.2 & 16.5 & 19.2 \\
0 & 31.6$_{\pm 0.9}$ & 25.3$_{\pm 1.0}$ & 23.7$_{\pm 2.1}$ & 16.8$_{\pm 2.8}$ & 19.8$_{\pm 1.3}$         \\
10 & 33.4$_{\pm 0.9}$ & 28.3$_{\pm 1.2}$ & 28.2$_{\pm 1.0}$ & 21.4$_{\pm 1.1}$ & 23.1$_{\pm 0.9}$\\
50 & \textbf{36.7$_{\pm 1.2}$} & 33.2$_{\pm 1.1}$ & \textbf{35.1$_{\pm 1.2}$} & 32.9$_{\pm 0.8}$ & \textbf{34.6$_{\pm 1.2}$}\\
100 & \textbf{38.4$_{\pm 1.1}$} & \textbf{37.3$_{\pm 0.9}$} & \textbf{36.8$_{\pm 1.5}$} & \textbf{38.2$_{\pm 0.9}$} & \textbf{36.1$_{\pm 1.0}$}\\
500 & \textbf{38.6$_{\pm 1.3}$} & \textbf{37.4$_{\pm 0.6}$} & \textbf{36.9$_{\pm 0.9}$} & \textbf{39.0$_{\pm 1.2}$} & \textbf{36.4$_{\pm 1.1}$}\\
1000 & \textbf{38.8$_{\pm 1.3}$} & \textbf{37.8$_{\pm 1.2}$} & \textbf{37.0$_{\pm 1.1}$} & \textbf{39.3$_{\pm 0.8}$} & \textbf{36.7$_{\pm 1.1}$}\\
\end{tabular}
}
\caption{Average F1 score when we vary the additional number of examples we augment from the existing safety rules in our \dapt(cosine) method}
\label{size}
\end{table*}

\end{document}